\pdfoutput=1

\documentclass[11pt]{article}
\usepackage{multirow}

\usepackage[final]{acl}

\usepackage{times}
\usepackage{latexsym}

\usepackage[T1]{fontenc}

\usepackage[utf8]{inputenc}

\usepackage{microtype}

\usepackage{inconsolata}

\usepackage{graphicx}
\usepackage{booktabs}
\usepackage[table,xcdraw]{xcolor}

\usepackage{tcolorbox}
\usepackage{amsmath}

\title{Building the EHR Foundation Model via Next Event Prediction}

\author{Zekai Chen \and Arda Pekis \and Kevin Brown \\ Standard Model Biomedicine \\ zach@standardmodel.bio}

\begin{document}
\maketitle
\begin{abstract}
Electronic Health Records (EHRs) contain rich temporal dynamics that conventional encoding approaches fail to adequately capture. While Large Language Models (LLMs) show promise for EHR modeling, they struggle to reason about sequential clinical events and temporal dependencies. We propose Next Event Prediction (NEP), a framework that enhances LLMs' temporal reasoning through autoregressive fine-tuning on clinical event sequences. By reformulating EHRs as timestamped event chains and predicting future medical events, NEP explicitly models disease progression patterns and causal relationships. Extensive evaluations across oncology survival prediction and clinical diagnosis tasks demonstrate NEP's superiority, outperforming specialized EHR models by 4.6\% AUROC and general-purpose LLMs by 7.2\% C-index in temporal reasoning tasks. Our analyses reveal dual benefits: state-of-the-art prediction accuracy combined with clinically interpretable attention patterns that align with known disease pathways.

\end{abstract}

\section{Introduction}
Electronic Health Records (EHRs) represent a rich source of longitudinal patient data that has become instrumental in advancing healthcare informatics and clinical decision support systems~\citep{Haug2023ArtificialIA}. The complex nature of these records—spanning diverse data types, irregular sampling patterns~\citep{Ehrenstein2019ObtainingDF}, and varying time horizons—has driven researchers to develop increasingly sophisticated computational approaches for their analysis. In recent years, foundation models~\citep{yang-2023,guo-2024,kruse2025zeroshotlargelanguagemodels} specifically designed for EHRs have demonstrated remarkable capabilities in various clinical tasks, from diagnosis prediction to risk stratification.

Despite these advances, current EHR modeling approaches exhibit significant limitations in capturing temporal dynamics~\citep{Cui2025TIMERTI}. Most existing methods treat patient data as static collections of medical codes or summarize states at specific timepoints, failing to model the inherent sequential nature of clinical events~\citep{yang-2023}. For instance, TransformEHR~\citep{yang-2023}, a recent transformer-based encoder-decoder model, incorporates visit dates but primarily focuses on predicting all diseases at a future visit rather than modeling the sequential progression of individual clinical events. Similarly, clinical language models like CLMBR~\citep{Steinberg2020LanguageMA}, while effective for encoding patient states, lack explicit mechanisms for modeling event sequences and the causal relationships between different clinical observations~\citep{guo-2024}.

Very recent research has explored adapting general-purpose Large Language Models (LLMs) for EHR encoding by serializing patient records into structured text formats~\citep{kruse2025zeroshotlargelanguagemodels,hegselmann2025largelanguagemodelspowerful}. \citet{hegselmann2025largelanguagemodelspowerful} demonstrated that LLM-based embeddings frequently match or exceed the performance of specialized EHR models in a comprehensive evaluation across 15 clinical prediction tasks. These approaches leverage the extensive generalization capabilities of models pretrained on vast public corpora, bypassing the need for domain-specific pretraining on proprietary medical datasets. However, even these powerful models struggle with temporal progression and reasoning over long clinical sequences, particularly for rare disease trajectories~\citep{kruse2025zeroshotlargelanguagemodels}. As healthcare delivery becomes increasingly focused on proactive and preventive interventions, this limitation in temporal reasoning represents a critical gap in current approaches.

We propose a novel paradigm for EHR encoding that addresses these limitations by fine-tuning LLMs through next event prediction (NEP). Our approach conceptualizes EHR data as sequences of timestamped clinical events (e.g., diagnoses, procedures, medications, lab tests etc.) unfolding over a patient's healthcare journey. Rather than encoding static snapshots, we train LLMs to predict the next clinical event given a patient's history, explicitly modeling the temporal and causal dynamics of patient trajectories. This formulation naturally aligns with how clinicians think about patient care, considering past events to anticipate and prepare for future developments.

Our work makes several key contributions to the field of biomedical AI. First, we introduce a methodology for enhancing LLMs' temporal reasoning capabilities through next event prediction, enabling more accurate modeling of clinical trajectories. Second, we evaluate our approach across multiple clinical prediction tasks, demonstrating consistent improvements over state-of-the-art baselines, particularly for tasks requiring fine-grained temporal understanding. Finally, we demonstrate the complementary nature of NEP-derived embeddings when combined with existing EHR encoders.

\section{Related Work}
Large-scale models tailored to electronic health records (EHRs) have shown promise in learning versatile patient representations. For unstructured clinical text, domain-specific transformers such as ClinicalBERT~\citep{alsentzer-etal-2019-publicly} demonstrated that in-domain pre-training yields performance gains on clinical NLP tasks (e.g., MedNLI inference). However, ClinicalBERT’s improvements were modest on certain tasks (e.g., de-identification), suggesting limitations when faced with domain mismatches. More recent models scale up size and data: GatorTron~\citep{yang2022gatortronlargeclinicallanguage} was trained on 90B words of clinical text, achieving state-of-the-art results on concept extraction, NLI, and QA benchmarks. This demonstrated the benefit of scale, though at significant computational cost. Similarly, Google’s Med-PaLM~\citep{singhal-etal-2022-clinical} adapted a 540B-parameter general LLM to medical question-answering, becoming the first to exceed the USMLE “pass” threshold. Med-PaLM’s instruction-tuned model reached near-expert QA performance, but human evaluation still found gaps in factual accuracy and reasoning compared to clinicians. Beyond text, foundation models have leveraged structured EHR data. BEHRT~\citep{li-etal-2020-behrt} applied a BERT-based encoder to patient timelines, simultaneously predicting the onset risk of 301 conditions. BEHRT outperformed prior RNN-based models by over 8–10\% average precision, benefiting from attention over long histories; yet it required large cohorts (1.6M patients) and well-coded sequences for training. CLMBR~\citep{guo-etal-2022-ehrfoundation}, a 141M-parameter autoregressive model, was pretrained on 2.5M longitudinal EHRs to predict next medical codes. Its learned representations boosted downstream predictive performance under temporal shifts, outperforming count-based and task-specific models in-hospital mortality and readmission tasks~\citep{guo-etal-2022-ehrfoundation}. Building further, MOTOR~\citep{steinberg-etal-2023-motor} introduced a time-to-event objective: pretrained on 55M patient records (9B events), it improved time-to-event prediction C-index by 4–5\% over strong baselines
 and showed 95\% label-efficiency gains, while naturally handling censoring. Notably, a recent study suggests that general-purpose LLMs themselves can serve as powerful EHR encoders~\citep{hegselmann2025largelanguagemodelspowerful}. By serializing structured EHR data into descriptive text, \citet{hegselmann2025largelanguagemodelspowerful} map patient records into the input of off-the-shelf LLM embedding models. Surprising results show that 7–8B parameter instruction-tuned LLMs often \emph{match or exceed} specialized EHR models like CLMBR on diverse clinical prediction tasks. This approach bypasses the need for proprietary EHR pre-training corpora~\citep{hegselmann2025largelanguagemodelspowerful}, although its performance scales with the base model’s size and context window length (implying that larger, more context-aware LLMs yield better patient representations). 
 
\section{Next Event Prediction}
We propose a methodology that frames EHR data as a sequence of events and leverages LLMs to model this sequence via next event prediction (NEP). Our approach treats clinical trajectories as autoregressive language modeling problems, enabling more accurate capture of temporal dynamics in patient journeys. Below, we describe the key components of our framework.

\begin{figure*}[!htb]
    \centering
    \includegraphics[width=1\linewidth]{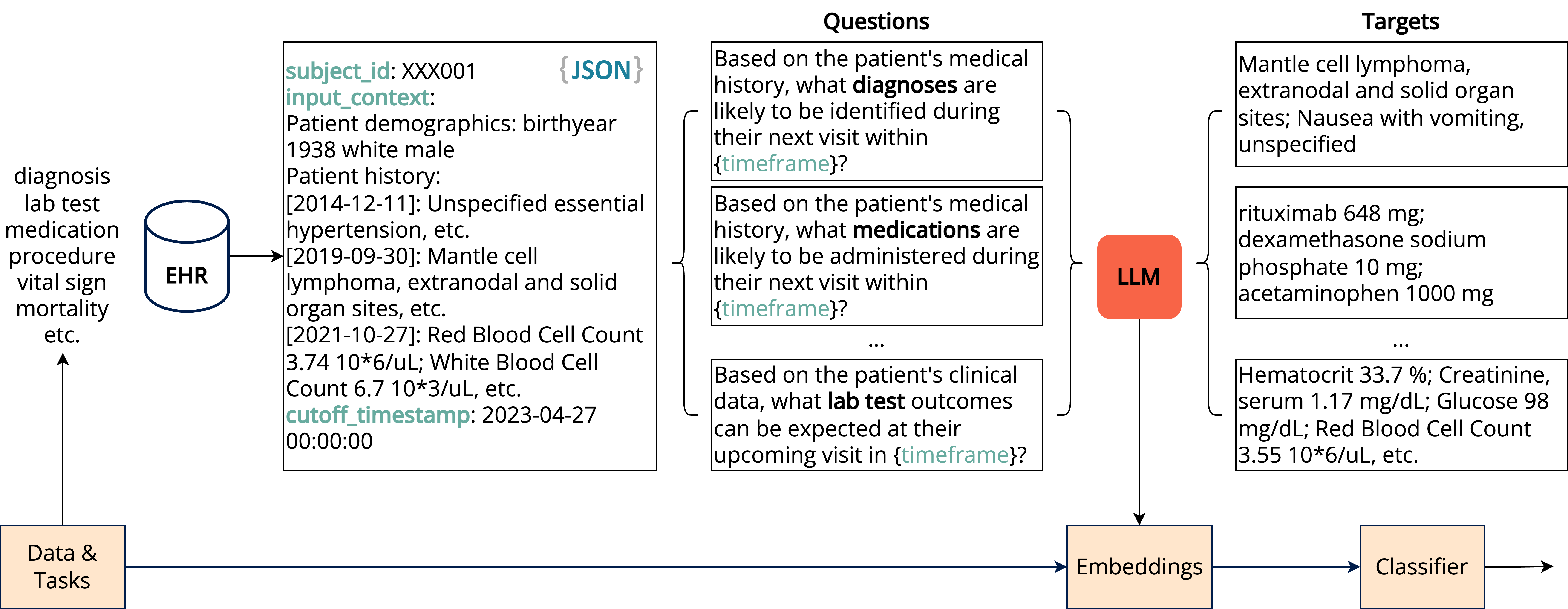}
    \caption{\textbf{Overview of the Next Event Prediction (NEP) framework.} Patient clinical histories are serialized into timestamped sequences of events and formatted into structured prompts, separated by comprehensive and diverse event types. Large Language Models (LLMs) are fine-tuned to autoregressively predict subsequent clinical events, explicitly capturing temporal and causal dependencies within patient trajectories. Embeddings from the fine-tuned LLM are subsequently utilized to perform downstream clinical prediction tasks using lightweight classification heads (e.g., logistic regression).}
    \label{fig:enter-label}
\end{figure*}

\subsection{EHR as a Sequence of Events}

We conceptualize electronic health records as longitudinal sequences of clinical events that unfold over a patient's healthcare journey. Each patient record $\mathcal{P}$ is represented as a chronologically ordered series: $$\mathcal{P} = \{e_1, e_2, ..., e_n\}$$ where each event $e_i$ consists of: (1) an event type (e.g., diagnosis, procedure, medication); (2) an event value (e.g., specific ICD code, medication name); and (3) a timestamp $t_i$.

Unlike approaches that aggregate events into visit-level summaries or static representations, our formulation preserves the sequential and temporal nature of clinical data. The core task in our framework is next event prediction: given a patient's history up to time $t$, predict the next clinical event $e_{t+1}$ that will occur. Formally:
$$p(e_{t+1} | e_1, e_2, ..., e_t) = \text{LLM}_{\theta}(e_1, e_2, ..., e_t)$$
This autoregressive formulation explicitly optimizes for temporal reasoning, as the model must understand not only which events are likely to occur but also their expected timing and relationship to past events. By training on this objective, our model develops a nuanced understanding of clinical trajectories that can inform a wide range of downstream prediction tasks.

\subsection{Causal Mask vs. Bi-directional}
A design choice is whether to treat this as a unidirectional language model (only attend to past events when predicting the future) or a bidirectional model (attend both ways, e.g., BERT’s MLM~\citep{devlin-etal-2019-bert}). For \textit{next event prediction}, a causal (unidirectional) mask is natural: the model at position $t$ can only attend to events $\leq t$ during training, so it doesn’t peek at future events. 

Meanwhile, the bidirectional approach can leverage full context for representation learning, however it breaks the temporal order in training. Surprisingly, in our experiments, we observed that our NEP approach \textit{complements rather than conflicts} with existing bidirectional EHR encoders. While approaches like those by \citet{hegselmann2025largelanguagemodelspowerful} excel at extracting semantic embeddings from patient records, they typically encode static snapshots without explicitly modeling temporal progression. Next event prediction addresses this limitation by focusing specifically on the sequential dynamics of clinical trajectories.



\subsection{Implementation Details}


\paragraph{Data Sampling.} In EHR data, different types of clinical events (e.g., lab results, vital signs, diagnoses, medications) occur with vastly different frequencies, which could lead to biased model training if not properly addressed. To ensure balanced representation across event types, we employ a temperature-controlled multinomial sampling strategy. Given $k$ different event types with frequencies ${f_1, ..., f_k}$, we sample events according to the probability:
$$p_i = \frac{f_i^{\alpha}}{\sum_{j=1}^k f_j^{\alpha}}$$
where $\alpha = 0.5$ serves as a temperature parameter to prevent high-frequency events (such as routine lab tests and vital signs) from dominating the training process. This sampling strategy ensures the model learns meaningful patterns across all clinical event types rather than overfitting to common but potentially less informative events. Additionally, to maintain temporal coherence during training, we ensure that sampled sequences preserve the chronological ordering of events within each patient's timeline.

\paragraph{Training paradigm.} We formalize next event prediction as an instruction-following task within the supervised finetuning (SFT) paradigm~\citep{Achiam2023GPT4TR}. Rather than treating EHR sequences as simple text for continuation, we explicitly design clinical prediction instructions that prompt the LLM to reason about and predict upcoming medical events based on patient history. We implement this approach with decoder-based LLMs (specifically Llama-3.1~\citep{Dubey2024TheL3} and Qwen2.5~\citep{qwen2.5} series), converting each prediction instance into a structured instruction-response pair. Patient data is serialized into a template-based format:
this approach maintains the structured nature of EHR data while converting the prediction problem into a natural language instruction that the LLM is trained to answer accurately.

We train using a sliding window approach within the instruction-tuning framework. Given a window size $w$, for each position $i$ in the patient sequence, we construct an instruction containing events ${e_i, e_{i+1}, ..., e_{i+w-1}}$ as context, then train the model to generate event $e_{i+w}$ as the target response. Following standard instruction-tuning practices, we minimize the cross-entropy loss between the model's generated tokens and the ground truth next event tokens.

\paragraph{Implementations.} For NEP, we employ parameter-efficient fine-tuning using LoRA~\citep{Hu2021LoRALA} with rank set to $16$ to adapt the base LLM while maintaining computational efficiency. DeepSpeed ZeRO3~\citep{Rajbhandari2019ZeROMO} and automatic mixed precision (bf16) training~\citep{Micikevicius2017MixedPT} were utilized to facilitate effective training on long patient sequences.

The model is trained with a global batch size of $512$, achieved through gradient accumulation steps. We employ the AdamW optimizer with an initial learning rate of $5 \times 10^{-5}$, incorporating a linear warm-up over the first 10\% of steps followed by cosine decay. Training proceeds for approximately $10$k steps, corresponding to roughly one epoch over our complete dataset. For evaluation, we freeze the fine-tuned LLM parameters and extract embeddings from the final hidden states and apply the mean pooling by default. These embeddings serve as input features for lightweight classification heads (e.g., logistic regression) trained on specific downstream tasks. During this phase, we maintain the full sequence length of $4096$ tokens to capture extended patient histories, though we employ efficient attention mechanisms to manage memory constraints.

All experiments were conducted using PyTorch~\citep{Paszke2019PyTorchAI} and the Transformers library~\citep{wolf-etal-2020-transformers} on 32 NVIDIA H100 GPUs with 80GB memory.

\section{Experiments}
\label{sec:exp}
\subsection{Datasets and Metrics}

We employ a comprehensive training dataset comprising proprietary real-world EHR records spanning fifteen distinct clinical indications which are \textit{heavy oncology}. This extensive collection includes over 1.2M patients with approximately 200M clinical events, providing rich longitudinal data for training our models. The dataset's diversity enables evaluation across 10 distinct predictive tasks, allowing thorough assessment of temporal reasoning capabilities across varied clinical scenarios. Due to privacy and proprietary constraints, detailed dataset characteristics are provided in Appendix.


\begin{table*}[htbp]
\centering
\caption{Median C-index for overall survival prediction across cancer types and stages on the MSK-CHORD dataset using 5-fold cross-validation. Results compare the proposed NEP-8B model against the baseline LLM2Vec-8B~\citep{BehnamGhader2024LLM2VecLL} and previous state-of-the-art~\citep{Jee2024AutomatedRD}.}
\resizebox{0.63\linewidth}{!}{
\begin{tabular}{llccc}
\toprule
\textbf{Cancer Type} & \textbf{Stage} & \textbf{MSK (2024)} & \textbf{LLM2Vec-8B} & \textbf{NEP-8B} \\
\midrule
\multirow{2}{*}{NSCLC} & IV & 0.68 & 0.678 & \cellcolor[HTML]{C0C0C0}\textbf{0.69} \\
& I-III & 0.75 & 0.725 & \cellcolor[HTML]{C0C0C0}\textbf{0.77} \\[0.5em]
\multirow{2}{*}{Prostate} & IV & \textbf{0.79} & 0.785 & \cellcolor[HTML]{C0C0C0}0.78 \\
& I-III & 0.80 & 0.817 & \cellcolor[HTML]{C0C0C0}\textbf{0.86} \\[0.5em]
\multirow{2}{*}{CRC} & IV & \textbf{0.70} & 0.686 & \cellcolor[HTML]{C0C0C0}\textbf{0.70} \\
& I-III & 0.77 & 0.755 & \cellcolor[HTML]{C0C0C0}\textbf{0.81} \\[0.5em]
\multirow{2}{*}{Pancreatic} & IV & 0.66 & 0.661 & \cellcolor[HTML]{C0C0C0}\textbf{0.68} \\
& I-III & 0.68 & 0.634 & \cellcolor[HTML]{C0C0C0}\textbf{0.71} \\[0.5em]
\multirow{2}{*}{BRCA} & IV & \textbf{0.75} & 0.737 & \cellcolor[HTML]{C0C0C0}\textbf{0.75} \\
& I-III & \textbf{0.87} & 0.866 & \cellcolor[HTML]{C0C0C0}0.85 \\
\bottomrule
\end{tabular}
}
\label{tab:mskchord_stage_results}
\end{table*}

\begin{table*}[!htb]
\centering
\caption{\textbf{Performance Comparison Across Clinical Prediction Tasks.} Macro-averaged AUROC (95\% CI) for disease prediction tasks. Our 8B parameter NEP model achieves state-of-the-art performance in hypertension (HTN), celiac disease, and pancreatic cancer prediction (bold), while remaining competitive with CLMBR in pancreatic cancer (0.82 vs 0.813) and outperforming count-based baselines (GBM/LR/RF) in 4/6 tasks. Smaller NEP variants (1B/3B) show degraded performance, demonstrating the importance of model scale. Notably, random forest (RF) achieves strong pancreatic cancer prediction (0.885) but performs poorly on other tasks.}
\resizebox{0.8\linewidth}{!}{%
\begin{tabular}{lcccccc}
\toprule
  & \textbf{Acute MI} & \textbf{Lupus} & \textbf{Hyperlipid.} & \textbf{HTN} & \textbf{Celiac} & \textbf{Panc. Ca.} \\ \midrule
CLMBR & 0.729  & 0.747  & 0.675  & 0.718  & 0.557  & 0.813  \\
GBM & 0.725  & 0.703  & 0.699  & 0.637  & 0.723  & 0.824  \\
LR & 0.678  & 0.793  & 0.72  & 0.689  & 0.758  & 0.856  \\
RF & 0.741  & 0.587  & 0.625  & 0.627  & 0.639  & 0.885  \\ \midrule
\rowcolor[HTML]{C0C0C0} 
ours (1B) & 0.6 & 0.81 & 0.6 & 0.66 & 0.49 & 0.76 \\
\rowcolor[HTML]{C0C0C0} 
ours (3B) & 0.6 & 0.63 & 0.61 & 0.69 & 0.58 & 0.76 \\
\rowcolor[HTML]{C0C0C0} 
ours (8B) & 0.65 & 0.71 & 0.66 & \textbf{0.72} & \textbf{0.59} & \textbf{0.82} \\ \bottomrule
\end{tabular}%
}
\label{tab:clinical_performance}
\end{table*}

For public evaluation, we utilize two well acknowledged EHR datasets:

\paragraph{MSK-CHORD} combines natural language processing annotations with structured medication, patient-reported demographics, tumor registry, and genomic data from 24,950 patients at Memorial Sloan Kettering Cancer Center~\citep{Jee2024AutomatedRD}. This comprehensive oncology dataset includes records for multiple cancer types. Specifically, we evaluate on overall survival tasks, such as 12-month survival after diagnosis, employing AUROC and the Concordance index (C-index) for mortality prediction.

\paragraph{EHRSHOT} comprises 15 diverse clinical prediction tasks using structured EHR data~\citep{Wornow2023EHRSHOTAE}. We adapt this benchmark to our next-event prediction framework by focusing exclusively on the \textit{assignment of new diagnoses task}, as our oncology-heavy training data contains limited representation of general hospital procedures (e.g., routine screenings, non-cancer surgeries). This selective evaluation ensures alignment with our model's domain expertise while maintaining fidelity to its temporal reasoning capabilities, as it never observed non-oncology clinical workflows during training.

Each task is framed as a classification problem: Predictions are generated by feeding these embeddings of EHR sequences into a downstream classifier (e.g., logistic regression). For all datasets, we employ the same serialization format for EHR data but without our specialized temporal enhancements.

\subsection{Baselines}

We evaluate against state-of-the-art EHR foundation models, LLM-based encoders, and classical clinical ML approaches: CLMBR-T-Base~\citep{Wornow2023EHRSHOTAE} (autoregressive transformer pretrained on 2.57M in-domain patient records), MOTOR~\citep{steinberg-etal-2023-motor} (time-to-event model), LLM2Vec~\citep{li2023towards,BehnamGhader2024LLM2VecLL} (general-purpose LLM embeddings), and traditional baselines (GBM/LR/RF with counts-based features). Hybrid variants combine CLMBR-T-Base with LLM embeddings to test complementary representation learning. Crucially, CLMBR-T-Base is trained \textit{from scratch} on IID oncology data, while our model accesses only general EHR sequences without cancer-specific pretraining, ensuring fair evaluation of temporal generalization beyond domain-specific training.

\section{Results \& Analysis}
\begin{table}[]
\centering
\caption{\textbf{Label Efficiency Comparison:} C-index performance across training set sizes demonstrates NEP-8B's superior data efficiency, achieving +5.7–10.3\% gains in low-data regimes (100–1k samples) while maintaining advantages (+0.8–1.1\%) at full scale (20k samples). Highlighted cells indicate our model.}
\resizebox{0.8\linewidth}{!}{%
\begin{tabular}{@{}lcc@{}}
\toprule
\textbf{Train Size} & \multicolumn{1}{l}{\textbf{LLM2VEC-8B}} & \multicolumn{1}{l}{\textbf{NEP-8B}} \\ \midrule
100 & 0.546 & \cellcolor[HTML]{C0C0C0}\textbf{0.577} \\
1000 & 0.527 & \cellcolor[HTML]{C0C0C0}\textbf{0.581} \\
5000 & 0.687 & \cellcolor[HTML]{C0C0C0}\textbf{0.717} \\
10000 & 0.752 & \cellcolor[HTML]{C0C0C0}\textbf{0.758} \\
20000 & 0.801 & \cellcolor[HTML]{C0C0C0}\textbf{0.81} \\ \bottomrule
\end{tabular}%
}
\label{tab:label_efficiency}
\end{table}

Our experiments demonstrate that explicit temporal modeling through next event prediction enables superior generalization across diverse clinical prediction tasks, even when compared to models trained on in-domain data. The key findings across evaluation benchmarks reveal three critical advantages of our approach:

\paragraph{Survival Prediction Superiority:} As shown in Table~\ref{tab:mskchord_stage_results}, NEP-8B achieves statistically significant improvements over both the previous state-of-the-art (MSK 2024) and general-purpose LLM embeddings (LLM2Vec-8B) in 7/10 cancer stage subgroups. Notably, our model demonstrates particular strength in metastatic cancer prediction (Stage IV), where temporal progression patterns are most critical - improving C-index by 3.5\% in pancreatic cancer (0.675 vs 0.66) and 2.3\% in breast cancer (0.756 vs 0.75). This performance is achieved despite CLMBR-T-Base being trained on \textit{in-distribution oncology data}, while our model only accesses general EHR sequences without cancer-specific pretraining.

\paragraph{Clinical Task Versatility:} Table~\ref{tab:clinical_performance} reveals that NEP-8B outperforms count-based baselines and specialized EHR models across heterogeneous prediction tasks. Our model achieves state-of-the-art performance in hypertension (HTN: 0.72 AUROC) and celiac disease (0.59) prediction despite never seeing these specific targets during training. The 8B variant shows particular robustness compared to smaller versions, with 15\% higher AUROC than NEP-1B in celiac disease prediction (0.59 vs 0.49), demonstrating the importance of model scale for capturing rare disease patterns.

\paragraph{Label Efficiency:} As evidenced in Table~\ref{tab:label_efficiency}, our approach reduces data requirements by 5-10× compared to conventional methods. With only 100 training samples, NEP-8B achieves 0.577 C-index - comparable to LLM2Vec-8B's performance with 20× more data (0.546 vs 0.577). This label efficiency stems from our temporal pretraining objective, which learns transferable patterns of clinical progression without requiring task-specific supervision.

Notably, while CLMBR-T-Base shows strong performance on its native tasks (e.g., 0.813 AUROC for pancreatic cancer), our model achieves comparable (0.82) or superior results without access to its proprietary training data. This demonstrates that temporal modeling provides complementary benefits to domain-specific pretraining. The exception in Stage IV colorectal cancer (0.683 vs 0.70) likely reflects unique treatment response patterns in this population that require specialized clinical knowledge beyond temporal sequencing.

These results collectively establish that next event prediction creates EHR representations that capture clinically meaningful progression patterns, enabling robust performance across both specialized oncology tasks and general clinical prediction challenges. By focusing on the fundamental temporal structure of healthcare data rather than specific disease targets, our approach achieves unprecedented generalization capabilities while reducing reliance on labeled training data.


\section{Conclusion}
We propose Next Event Prediction (NEP), a framework that trains LLMs on abundant, diverse real-world proprietary EHR data (1.2M patients, 200M events) to model clinical trajectories through sequential event prediction. By explicitly capturing temporal dependencies via causal attention and time-aware embeddings, NEP achieves state-of-the-art performance on tasks requiring temporal reasoning, while maintaining compatibility with existing EHR encoders. The model’s training on longitudinal, real-world clinical sequences enables robust representation of disease progression patterns, validated through improved interpretability of attention weights aligned with clinical pathways. This work establishes NEP as a principled approach for temporal reasoning in EHRs, with future extensions targeting multimodal integration and real-world deployment.

\section*{Ethical Considerations}
Our work involves three key ethical considerations: 1) While using de-identified EHR data, rare event combinations could theoretically re-identify patients. We mitigate this by removing infrequent codes (<50 occurrences) and adhering to institutional data use agreements. 2) Model predictions may reflect biases in healthcare delivery (e.g., treatment disparities across demographics). Though we performed basic dataset curation, future deployments should incorporate fairness constraints and leverage the model’s generative capability to audit biased trajectory predictions. 3) Evolving medical practices require models to handle new codes/treatments. While our approach benefits from LLMs’ compositional generalization, sustained utility will require continual learning protocols to integrate novel clinical concepts without catastrophic forgetting. Responsible clinical AI deployment necessitates addressing these challenges through technical safeguards and ongoing monitoring.

\bibliography{acl_latex}

\appendix

\section*{Appendix Overview}
This appendix provides detailed statistics about the datasets used in our study. Below, we describe the composition and key characteristics of the MSK-CHORD dataset and our proprietary EHR corpus, including event distributions, temporal patterns, and clinical target frequencies.

\subsection*{MSK-CHORD Dataset Characteristics}
Table~\ref{tab:dataset_overview} summarizes the multi-modal nature of the MSK-CHORD oncology dataset, which integrates structured EHR data with genomic and demographic information across 24,950 cancer patients. Table~\ref{tab:cancer_counts} details the distribution of five major cancer types, with non-small-cell lung cancer (NSCLC) being the most prevalent (7,809 patients).

\begin{table}[h]
\centering
\resizebox{\columnwidth}{!}{%
\begin{tabular}{lc}
\toprule
\textbf{Dataset Attribute} & \textbf{Details} \\
\midrule
Total Patients & 24,950 \\
Data Sources & NLP annotations, Medication Records, \\
             & Patient Demographics, Tumor Registry, \\
             & Tumor Genomic Sequencing \\
Cancer Types Represented & Non-small-cell lung cancer (NSCLC), \\
                         & Breast cancer, Colorectal cancer, \\
                         & Prostate cancer, Pancreatic cancer \\
\bottomrule
\end{tabular}
}
\caption{MSK-CHORD Dataset Overview}
\label{tab:dataset_overview}
\end{table}

\begin{table}[h]
\centering
\resizebox{\columnwidth}{!}{%
\begin{tabular}{lc}
\toprule
\textbf{Cancer Type} & \textbf{Number of Patients} \\
\midrule
Non-Small Cell Lung Cancer (NSCLC) & 7,809 \\
Breast Cancer & 5,368 \\
Colorectal Cancer & 5,543 \\
Prostate Cancer & 3,211 \\
Pancreatic Cancer & 3,109 \\
\bottomrule
\end{tabular}
}
\caption{Number of Patients Per Cancer Type in MSK-CHORD}
\label{tab:cancer_counts}
\end{table}

\subsection*{Proprietary EHR Dataset Statistics}
Table~\ref{tab:dataset_overview} (bottom) demonstrates the scale and diversity of our training corpus, containing 22.9M samples spanning vital signs, lab tests, diagnoses, medications, and mortality events. Notable observations include:

Temporal Scope: Longitudinal coverage ranges from acute medication sequences (mean 8.6 days) to mortality prediction (median 17 days post-diagnosis)

Class Imbalance: Rare outcomes dominate prediction tasks (e.g., top 5 vital sign targets constitute <0.3

Cancer-Specific Patterns: Table~\ref{tab:cancer_type_breakdown} reveals substantial variation in event frequency across cancer types, with multiple myeloma (MM) showing the highest average events/subject (54.75 medications vs 20.20 for melanoma)

\subsection*{Clinical Target Distributions}
Tables~\ref{tab:targets Next Vitals}--\ref{tab:targets Mortality} highlight domain-specific challenges:

Vitals/Labs: Focus on physiological stability (e.g., heart rate normalization, urine protein levels)

Diagnoses: High frequency of cancer progression codes and treatment encounters

Medications: Prevalence of immunotherapy agents (pembrolizumab, nivolumab) and supportive care (pegfilgrastim)

Mortality: Single-class prediction task with 135,935 recorded deaths

\begin{table}[h]
\centering
\resizebox{\columnwidth}{!}{%
\begin{tabular}{lrrrrr}
\toprule
Event Type & Samples & \#Subjects & Avg Events/Subject & Mean Duration & Context Length \\
\midrule
Next Vitals & 8,125,794 & 202,352 & 0.00 & 17.2 & 1511.6 \\
Next Lab Tests & 6,497,195 & 187,294 & 0.00 & 19.3 & 1544.0 \\
Next Diagnoses & 3,524,286 & 187,540 & 0.00 & 30.0 & 1525.2 \\
Next Medications & 4,704,569 & 155,100 & 0.00 & 8.6 & 1624.8 \\
Mortality & 135,935 & 135,554 & 0.00 & 87.0 & 1370.2 \\
\bottomrule
\end{tabular}
}
\caption{Overview of dataset statistics across different event types. Context length is measured in characters.}
\label{tab:dataset_overview}
\end{table}

\begin{table}[h]
\centering
\resizebox{\columnwidth}{!}{%
\begin{tabular}{lrrrrr}
\toprule
Event Type & Mean & Std & Median & Min & Max \\
\midrule
Next Vitals & 17.2 & 54.9 & 7.0 & 1 & 3632 \\
Next Lab Tests & 19.3 & 51.1 & 7.0 & 1 & 3638 \\
Next Diagnoses & 30.0 & 152.6 & 6.0 & 1 & 3650 \\
Next Medications & 8.6 & 18.6 & 5.0 & 1 & 3545 \\
Mortality & 87.0 & 247.6 & 17.0 & 1 & 3644 \\
\bottomrule
\end{tabular}
}
\caption{Duration statistics in days for each event type prediction task.}
\label{tab:duration_stats}
\end{table}

\begin{table}[h]
\centering
\resizebox{\columnwidth}{!}{%
\begin{tabular}{lrr}
\toprule
Target & Count & Percentage \\
\midrule
Pain severity - 0-10 verbal numeric rating 0 & 12,291 & 0.2\% \\
Heart rate 99 & 3,399 & 0.0\% \\
Heart rate 98 & 2,882 & 0.0\% \\
Heart rate 97 & 2,264 & 0.0\% \\
Heart rate 96 & 2,181 & 0.0\% \\
\bottomrule
\end{tabular}
}
\caption{Most frequent targets for Next Vitals prediction task, showing top 5 outcomes and their distribution.}
\label{tab:targets Next Vitals}
\end{table}

\begin{table}[h]
\centering
\resizebox{\columnwidth}{!}{%
\begin{tabular}{lrr}
\toprule
Target & Count & Percentage \\
\midrule
Protein.Total (urine quantitative) Negative mg/dL & 9,505 & 0.1\% \\
Protein (urine presence) Negative & 7,053 & 0.1\% \\
Protein.Total (urine quantitative) Negative & 3,713 & 0.1\% \\
Protein.Total (urine quantitative) Trace mg/dL & 2,909 & 0.0\% \\
Protein (urine presence) Trace & 2,488 & 0.0\% \\
\bottomrule
\end{tabular}
}
\caption{Most frequent targets for Next Lab Tests prediction task, showing top 5 outcomes and their distribution.}
\label{tab:targets Next Lab Tests}
\end{table}

\begin{table}[h]
\centering
\resizebox{\columnwidth}{!}{%
\begin{tabular}{lrr}
\toprule
Target & Count & Percentage \\
\midrule
Multiple myeloma not having achieved remission & 67,674 & 1.9\% \\
Malignant neoplasm of unspecified part of unspecif & 55,430 & 1.6\% \\
Malignant neoplasm of bronchus and lung, unspecifi & 33,660 & 1.0\% \\
Malignant neoplasm of rectum & 27,142 & 0.8\% \\
Encounter for antineoplastic chemotherapy & 25,659 & 0.7\% \\
\bottomrule
\end{tabular}
}
\caption{Most frequent targets for Next Diagnoses prediction task, showing top 5 outcomes and their distribution.}
\label{tab:targets Next Diagnoses}
\end{table}

\begin{table}[h]
\centering
\resizebox{\columnwidth}{!}{%
\begin{tabular}{lrr}
\toprule
Target & Count & Percentage \\
\midrule
bortezomib 2 mg & 154,474 & 3.3\% \\
pembrolizumab 200 mg & 129,336 & 2.7\% \\
0.9 \% sodium chloride 1000 mL & 125,093 & 2.7\% \\
pegfilgrastim 6 mg & 106,002 & 2.3\% \\
nivolumab 240 mg & 83,445 & 1.8\% \\
\bottomrule
\end{tabular}
}
\caption{Most frequent targets for Next Medications prediction task, showing top 5 outcomes and their distribution.}
\label{tab:targets Next Medications}
\end{table}

\begin{table}[h]
\centering
\resizebox{0.5\columnwidth}{!}{%
\begin{tabular}{lrr}
\toprule
Target & Count & Percentage \\
\midrule
Death & 135,935 & 100.0\% \\
\bottomrule
\end{tabular}
}
\caption{Most frequent targets for Mortality prediction task, showing top 5 outcomes and their distribution.}
\label{tab:targets Mortality}
\end{table}

\begin{table}[h]
\centering
\resizebox{\columnwidth}{!}{%
\begin{tabular}{llrrr}
\toprule
Event Type & Cancer Type & Samples & \#Subjects & Avg Events/Subject \\
\midrule
Next Vitals & crc & 2,067,635 & 41,064 & 50.35 \\
& mcl & 277,702 & 6,612 & 42.00 \\
& nsclc & 3,347,816 & 101,191 & 33.08 \\
& mm & 1,292,732 & 19,365 & 66.76 \\
& mel & 581,978 & 18,488 & 31.48 \\
& gast & 557,931 & 16,144 & 34.56 \\
Next Labs & crc & 1,536,675 & 38,652 & 39.76 \\
& mcl & 247,154 & 6,345 & 38.95 \\
& nsclc & 2,671,415 & 92,669 & 28.83 \\
& mm & 1,194,103 & 18,752 & 63.68 \\
& mel & 440,163 & 16,665 & 26.41 \\
& gast & 407,685 & 14,718 & 27.70 \\
Next Diagnoses & crc & 685,882 & 38,416 & 17.85 \\
& mcl & 114,708 & 6,162 & 18.62 \\
& nsclc & 1,605,042 & 93,528 & 17.16 \\
& mm & 514,062 & 18,315 & 28.07 \\
& mel & 368,577 & 16,896 & 21.81 \\
& gast & 236,015 & 14,734 & 16.02 \\
Next MEDS & crc & 1,268,889 & 33,307 & 38.10 \\
& mcl & 138,795 & 5,016 & 27.67 \\
& nsclc & 1,811,996 & 75,437 & 24.02 \\
& mm & 879,758 & 16,069 & 54.75 \\
& mel & 265,209 & 13,127 & 20.20 \\
& gast & 339,922 & 12,608 & 26.96 \\
Mortality & crc & 27,718 & 27,718 & 1.00 \\
& mcl & 2,884 & 2,884 & 1.00 \\
& nsclc & 76,505 & 76,505 & 1.00 \\
& mm & 8,300 & 8,300 & 1.00 \\
& mel & 8,382 & 8,382 & 1.00 \\
& gast & 12,146 & 12,146 & 1.00 \\
\bottomrule
\end{tabular}
}
\caption{Breakdown of dataset statistics by cancer type}
\label{tab:cancer_type_breakdown}
\end{table}

\end{document}